\def\year{2021}\relax
\documentclass[letterpaper]{article} 
\usepackage{aaai21}  
\usepackage{times}  
\usepackage{helvet} 
\usepackage{courier}  
\usepackage[hyphens]{url}  
\usepackage{graphicx} 
\usepackage{natbib}  
\usepackage{caption} 
\usepackage{amsmath}
\usepackage{amssymb}
\usepackage{dsfont}
\usepackage{booktabs}
\usepackage{multirow}

\usepackage{subcaption}
\title{Deep Open Intent Classification with Adaptive Decision Boundary}

\author{
	Hanlei Zhang,\textsuperscript{\rm 1, 2}
	Hua Xu,\textsuperscript{\rm 1, 2}\thanks{Hua Xu is the corresponding author.}
	Ting-En Lin\textsuperscript{\rm 1, 2}\\
}
\affiliations{
	\textsuperscript{\rm 1}State Key Laboratory of Intelligent Technology and Systems, \\
	Department of Computer Science and Technology, 
	Tsinghua University, Beijing 100084, China,\\
	\textsuperscript{\rm 2} Beijing National Research Center for Information Science and Technology (BNRist), Beijing 100084, China\\
	zhang-hl20@mails.tsinghua.edu.cn,\quad xuhua@tsinghua.edu.cn, \quad ting-en.lte@alibaba-inc.com
}
\begin{document}
	
	\maketitle
	
	\begin{abstract}
		Open intent classification is a challenging task in dialogue systems. On the one hand, it should ensure the quality of known intent identification. On the other hand, it needs to detect the open (unknown) intent without prior knowledge. Current models are limited in finding the appropriate decision boundary to balance the performances of both known intents and the open intent. In this paper, we propose a post-processing method to learn the adaptive decision boundary (ADB) for open intent classification. We first utilize the labeled known intent samples to pre-train the model. Then, we automatically learn the adaptive spherical decision boundary for each known class with the aid of well-trained features. Specifically, we propose a new loss function to balance both the empirical risk and the open space risk. Our method does not need open intent samples and is free from modifying the model architecture. Moreover, our approach is surprisingly insensitive with less labeled data and fewer known intents. Extensive experiments on three benchmark datasets show that our method yields significant improvements compared with the state-of-the-art methods. The codes are released at \url{https://github.com/thuiar/Adaptive-Decision-Boundary}.
		
	\end{abstract}
	
	\section{Introduction}
\begin{figure}[t!]
		\centering  
		\includegraphics[width=0.95\columnwidth]{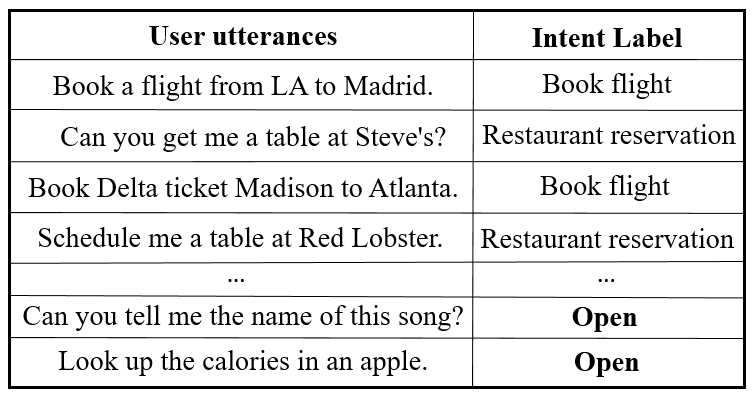}
		\caption{\label{example} An example of open intent classification. Book flight and Restaurant reservation are two known intents. We should identify them correctly while detecting the sentences with the open intent. }
	\end{figure}

	Identifying the user's open intent plays a significant role in dialogue systems. As shown in Figure~\ref{example}, we have two known intents for specific purposes, such as book flight and restaurant reservation. However, there are also utterances with irrelevant or unsupported intents that our system cannot handle. It is necessary to distinguish these utterances from the known intents as much as possible. On the one hand, effectively identifying the open intent can improve customer satisfaction by reducing false-positive error. On the other hand, we can use the open intent to discover potential user needs.
	
	We regard open intent classification as an (n+1)-class classification task as suggested in~\cite{Shu2017DOCDO,lin-xu-2019-deep}, and group open classes into the (n+1)$^{\text{th}}$ class . Our goal is to classify the n-class known intents into their corresponding classes correctly while identifying the (n+1)$^{\text{th}}$ class open intent. To solve this problem,~\citet{scheirer2013toward} propose the concept of open space risk as the measure of open classification.~\citet{fei-liu-2016-breaking} reduce the open space risk by learning the closed boundary of each positive class in the similarity space. However, they fail to capture high-level semantic concepts with SVM. 
	\begin{figure*}
		\centering
		\includegraphics[scale=.6]{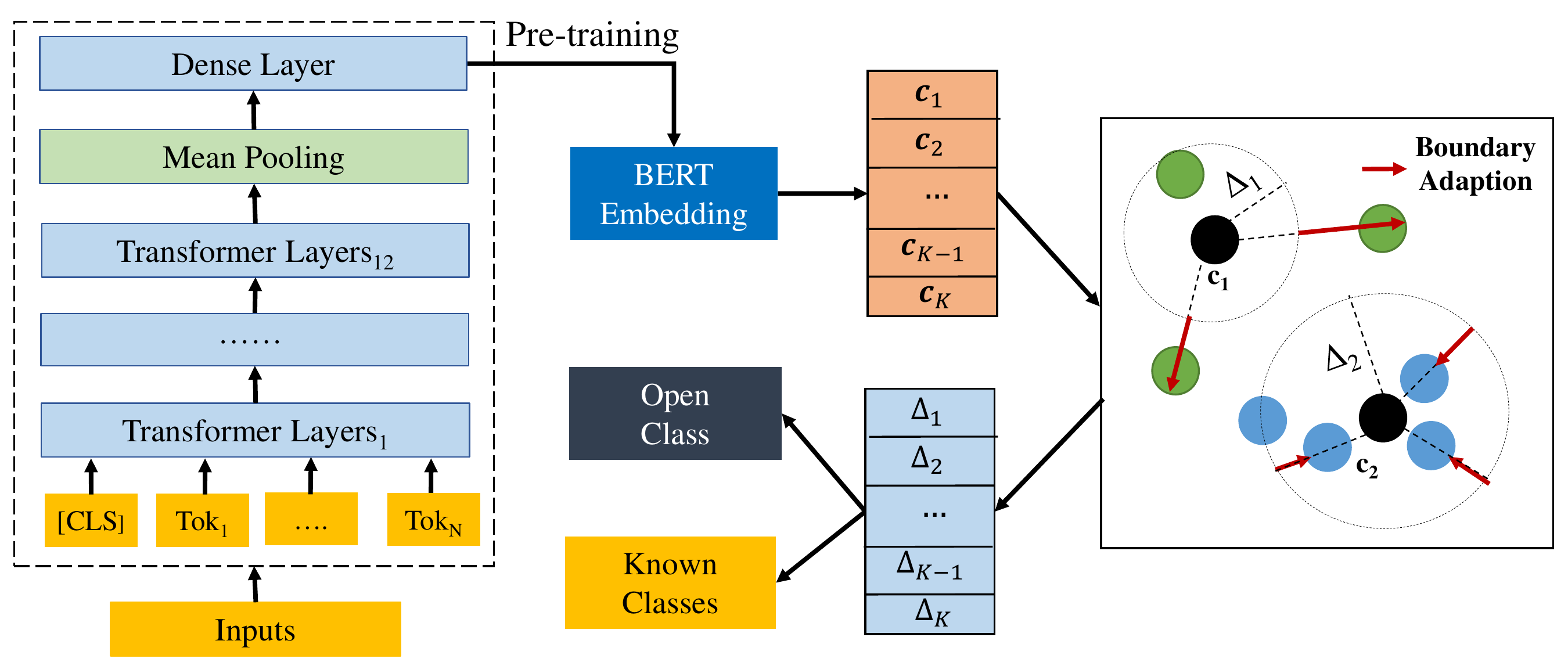}
		\caption{The model architecture of our approach. Firstly, we use BERT to extract intent features and pre-train the model with labeled samples. Then, we initialize the centroids  $\left\{\boldsymbol{c}_{i}\right\}_{i=1}^{K}$  and the radius of decision boundaries $\left\{\Delta_{i}\right\}_{i=1}^{K}$ for each known class. Next, we propose the boundary loss to learn tight decision boundaries adaptive to the known intent feature space. Finally, we perform open classification with the learned decision boundaries to both identify known classes and detect the open class.}
		\label{model}
	\end{figure*}~\citet{bendale2016towards} manage to reduce the open space risk through deep neural networks (DNNs), but need to sample open classes for selecting the core hyperparameters.~\citet{hendrycks17baseline} use the softmax probability as the confidence score, but also need to select the confidence threshold with negative samples.~\citet{Shu2017DOCDO} replace softmax with the sigmoid activation function, and calculate the confidence thresholds of each class based on statistics. However, the statistics-based thresholds can not learn the essential differences between known classes and the open class.~\citet{lin-xu-2019-deep} propose to learn the deep intent features with the margin loss and detect the unknown intent with local outlier factor~\cite{breunig2000lof}. However, it has no specific decision boundaries for distinguishing the open intent, and needs model architecture modification. 
	
	Most of the existing methods need to design specific classifiers for identifying the open class~\cite{bendale2016towards,Shu2017DOCDO,lin-xu-2019-deep}, and perform poorly with the common classifier~\cite{hendrycks17baseline}. Moreover, the performance of open classification largely depends on the  decision conditions. Most of these methods need negative samples for determining the suitable decision conditions~\cite{scheirer2013toward,fei-liu-2016-breaking,hendrycks17baseline,liang2018enhancing}. It is also a complicated and time-consuming process to manually select the optimal decision condition, which is not applicable in real scenarios. 
	
	To solve these problems, we use known intents as prior knowledge, and propose a novel post-processing method to learn the adaptive decision boundary (ADB) for open intent classification. As illustrated in Figure~\ref{model}, we first extract intent representations from BERT~\cite{devlin2019bert}. Then, we pre-train the model under the supervision of the softmax loss. We define centroids for each known class and suppose intent features of each known class are constrained in a closed ball area. Next, we aim to learn the radius of the spherical area to obtain the decision boundaries. Specifically, we initialize the boundary parameters with standard normal distribution and use a learnable activation function as a projection to get the radius of each decision boundary. 
	
	The suitable decision boundaries should satisfy two conditions. On the one hand, they should be broad enough to surround known intent samples as much as possible. On the other hand, they need to be tight enough to prevent the open intent samples from being identified as known intents. To address these issues, we propose a new loss function, which optimizes the boundary parameters by balancing both the open space risk and the empirical risk~\cite{scheirer2013toward}. The decision boundaries can automatically learn to adapt to the intent feature space with the boundary loss until balance. We find our post-processing method can learn discriminative decision boundaries for detecting the open intent even without modifying the original model architecture.
	
	We summarize our contribution as follows. Firstly, we propose a novel post-processing method for open classification, with no need for prior knowledge of the open intent. Secondly,  we propose a new loss function to automatically learn tight decision boundaries adaptive to the feature space. To the best of our knowledge, this is the first attempt to adopt deep neural networks to learn the adaptive decision boundaries for open classification. Thirdly, extensive experiments conducted on three challenging datasets show that our approach yields consistently better and more robust results compared with the state-of-the-art methods. 
	\begin{table*}[t!]\small
		\centering
		\begin{tabular}{@{\extracolsep{15pt}} ccccccc @{}}
			\toprule
			Dataset & Classes & \#Training & \#Validation & \#Test & Vocabulary Size & Length (max / mean) \\
			\midrule
			BANKING & 77 & 9,003 & 1,000 & 3,080 & 5,028 & 79 / 11.91\\
			OOS & 150 & 15,000 & 3,000 & 5,700 & 8,376 & 28 / 8.31 \\
			StackOverflow & 20 & 12,000 & 2,000 & 6,000 & 17,182 & 41 / 9.18 \\
			\bottomrule
		\end{tabular}
		\caption{ \label{data-stat-table} Statistics of BANKING, OOS and StackOverflow datasets. \# indicates the total number of sentences.}
	\end{table*}
	\section{The Proposed Approach}
	\subsection{Intent Representation}
	We use the BERT model to extract deep intent features. Given $i^{th}$ input sentence $\boldsymbol{s}_{i}$, we get all its token embeddings $[CLS, T_1, \cdots, T_N]$ $\in$ $\mathds R^{(N+1) \times H}$ from the last hidden layer of BERT. As suggested in~\cite{lin2020discovering}, we  perform mean-pooling on these token embeddings to synthesize the high-level semantic features in one sentence, and get the averaged representation $\boldsymbol{x}_{i} \in \mathds R^{H}$:
	\begin{align}
		\boldsymbol{x}_{i} = \text{mean-pooling}([CLS, T_1, \cdots, T_N]),
	\end{align}
	where $CLS$ is the vector for text classification, $N$ is the sequence length and $H$ is the hidden layer size. To further strengthen feature extraction capability, we feed $\boldsymbol{x}_{i}$ to a dense layer $h$ to get the intent representation $\boldsymbol{z}_i \in \mathds R^{D}$:
	\begin{align}
		\boldsymbol{z}_i=h(\boldsymbol{x}_i) = \sigma(W_h\boldsymbol{x}_{i}+b_h),
	\end{align}
	where $D$ is the dimension of the intent representation, $\sigma$ is a ReLU activation function, $W_h \in \mathds R^{H \times D}$ and $b_h \in \mathds R^{D}$ respectively denote the weights and the bias term of layer $h$. 
	
	\subsection{Pre-training}
	As the decision boundaries learn to adapt to the intent feature space, we need to learn intent representations at first. Due to lack of open intent samples, we use known intents as prior knowledge to pre-train the model. In order to reflect the effectiveness of the learned decision boundary, we  use the simple softmax loss $\mathcal{L}_{s}$ to learn the intent feature $\boldsymbol{z}_i$:
	\begin{align}
		\mathcal{L}_{s}=-\frac{1}{N}\sum_{i=1}^{N} \log\frac{\exp(\phi(\boldsymbol{z}_{i})^{y_{i}})}{\sum_{j=1}^{K}\exp(\phi(\boldsymbol{z}_{i})^{j})},
	\end{align}
	where $\phi(\cdot)$ is a linear classifier and $\phi(\cdot)^{j}$ are the output logits of the $j^{th}$ class. Then, we use the pre-trained model to extract intent features for learning decision boundaries. 
	
	\subsection{Adaptive Decision Boundary Learning}
	In this section, we propose our approach to learning the adaptive decision boundary (ADB) for open intent classification. First, we introduce the formulation of the decision boundary. Then, we propose our boundary learning strategy for optimization. Finally, we use the learned decision boundary to perform open classification.  
	
	\subsubsection{Decision Boundary Formulation}
	It has been shown the superiority of the spherical shape boundary for open classification~\cite{fei-liu-2016-breaking}. Compared with the half-space binary linear classifier~\cite{scholkopf2001estimating} or two parallel hyper-planes~\cite{scheirer2013toward}, the bounded spherical area greatly reduces the open space risk. Inspired by this, we aim to learn the decision boundary of each class constraining the known intents within a ball area.
	
	Let $S=\{(\boldsymbol{z}_i, y_{i}),\ldots, (\boldsymbol{z}_N, y_{N})\}$ be the known intent examples with their corresponding labels. $S_{k}$ denotes the set of examples labeled with class $k$. The centroid $\boldsymbol{c}_{k} \in \mathds  R^{D}$ is the mean vector of embedded samples in $S_{k}$:
	\begin{align}
		\boldsymbol{c}_{k}=\frac{1}{\left|S_{k}\right|} \sum_{\left(\boldsymbol{z}_i, y_{i}\right) \in S_{k}} \boldsymbol{z}_i,
		\label{center}
	\end{align}
	where $|S_{k}|$ denotes the number of examples in $S_{k}$. We define $\Delta_{k}$ as the radius of the decision boundary with respect to the centroid $\boldsymbol{c}_{k}$. For each known intent $\boldsymbol{z}_{i}$, we aim to satisfy the following constraints:
	\begin{align}
		\forall \boldsymbol{z}_{i} \in S_{k},  \left\|\boldsymbol{z}_{i}-\boldsymbol{c}_{k}\right\|_{2} \leq \Delta_{k},
	\end{align}
	where $\|\boldsymbol{z}_{i}-\boldsymbol{c}_{k}\|_{2}$ denotes the Euclidean distance between $\boldsymbol{z}_{i}$ and $\boldsymbol{c}_k$. That is, we hope examples belonging to class $k$ are constrained in the ball area with centroid $\boldsymbol{c}_{k}$ and radius $\Delta_{k}$. As radius $\Delta_{k}$ needs to be adaptive to the intent feature space, we use the deep neural network to optimize 
	the learnable boundary parameter $\widehat{\Delta_{k}} \in \mathds R$. As suggested in~\cite{tapaswi2019video}, we use $\operatorname{Softplus}$ activation function as the mapping between $\Delta_{k}$ and $\widehat{\Delta_{k}}$:
	\begin{align}
		\Delta_{k}=\log \left(1+e^{\widehat{\Delta_{k}}}\right).
		\label{delta}
	\end{align}
	
	The $\operatorname{Softplus}$ activation function has the following advantages. First, it is totally differentiable with different $\widehat{\Delta_{k}} \in \mathds R$. Second, it can ensure the learned radius $\Delta_{k}$ is above zero. Finally, it achieves linear characteristics like ReLU  and allows for bigger $\Delta_{k}$ if necessary.
	\begin{table*}[!t]\small
		\centering
		\begin{tabular}{@{\extracolsep{15pt}}clccccccccc}
			\toprule
			\centering
			&  & \multicolumn{2}{c}{BANKING} & \multicolumn{2}{c}{OOS} & \multicolumn{2}{c}{StackOverflow} \\
			
			\addlinespace[0.1cm]  \cline{3-4} \cline{5-6} \cline{7-8}  \addlinespace[0.1cm]
			& Methods &  Accuracy & F1-score   & Accuracy & F1-score & Accuracy & F1-score\\
			\midrule
			\multirow{6}{*}[1ex]{25\%}
			& MSP  & 43.67 & 50.09  & 47.02 & 47.62   & 28.67 & 37.85\\
			& DOC  & 56.99 & 58.03  & 74.97 & 66.37  & 42.74 & 47.73\\
			& OpenMax  & 49.94 & 54.14  & 68.50 & 61.99  & 40.28 & 45.98\\
			& DeepUnk  & 64.21 & 61.36  & 81.43 & 71.16  & 47.84 & 52.05 \\									
			& ADB   & \textbf{78.85} & \textbf{71.62}  & \textbf{87.59} & \textbf{77.19}   & \textbf{86.72} & \textbf{80.83}\\
			\midrule
			\midrule
			\multirow{6}{*}[1ex]{50\%}
			& MSP & 59.73 & 71.18  & 62.96 & 70.41  & 52.42 & 63.01\\
			& DOC  & 64.81 & 73.12  & 77.16 & 78.26    & 52.53 & 62.84 \\
			& OpenMax  & 65.31 & 74.24  & 80.11 & 80.56  & 60.35 & 68.18\\
			& DeepUnk  & 72.73 & 77.53  & 83.35 & 82.16  & 58.98 & 68.01 \\
			& ADB 	&\textbf{78.86}	&\textbf{80.90}		&\textbf{86.54}	&\textbf{85.05}	&\textbf{86.40}	&\textbf{85.83}\\
			\midrule
			\midrule
			\multirow{6}{*}[1ex]{ 75\%}
			& MSP  & 75.89 & 83.60  & 74.07 & 82.38   & 72.17& 77.95\\
			& DOC  & 76.77 & 83.34  & 78.73 & 83.59 & 68.91 & 75.06\\
			& OpenMax  & 77.45 & 84.07  & 76.80 & 73.16   & 74.42 & 79.78\\
			& DeepUnk  & 78.52 & 84.31  & 83.71 & 86.23  & 72.33 & 78.28\\
			& ADB 	&\textbf{81.08}	&\textbf{85.96}		&\textbf{86.32}	&\textbf{88.53}	&\textbf{82.78}	&\textbf{85.99}\\
			\bottomrule
		\end{tabular}
		\caption{ \label{results-main-1}  
			Results of open classification with different known class proportions (25\%, 50\% and 75\%) on BANKING, OOS and StackOverflow datasets. “Accuracy” and “F1-score” respectively denote the
			accuracy score and macro F1-score over all classes.
		}
	\end{table*}
	\subsubsection{Boundary Learning}
	The decision boundaries should be adaptive to the intent feature space to balance both empirical and open space risk~\cite{Bendale2015}. For example, if $\|\boldsymbol{z}_{i}-\boldsymbol{c}_{y_{i}}\|_{2} > \Delta_{y_{i}}$, the known intent samples are outside their corresponding decision boundaries, which may introduce more empirical risk. Therefore, the decision boundaries need to expand to contain more samples from known classes. If $\|\boldsymbol{z}_{i}-\boldsymbol{c}_{y_{i}}\|_{2} < \Delta_{y_{i}}$, though more known intent samples are likely to be identified with broader decision boundaries, it may introduce more open intent samples and increase the open space risk. Thus, we propose the boundary loss $\mathcal{L}_{b}$:
	\begin{equation}
	\begin{split}
	\mathcal{L}_{b}=\frac{1}{N}&\sum_{i=1}^{N}\left[\delta_{i}\left(\|\boldsymbol{z}_{i}-\boldsymbol{c}_{y_{i}}\|_{2}
	-\Delta_{y_{i}}\right)\right.\\
	&+\left. \left(1-\delta_{i}\right) \left(\Delta_{y_{i}}-\|\boldsymbol{z}_{i}-\boldsymbol{c}_{y_{i}}\|_{2}
	\right)\right],
	\end{split} 
	\end{equation}
	where $y_{i}$ is the label of the $i^{th}$ sample and $\delta_{i}$ is defined as:
	\begin{equation}
	\delta_{i}:=\left\{\begin{array}{ll}1, \text { if } & \|\boldsymbol{z}_{i}-\boldsymbol{c}_{y_{i}}\|_{2} > \Delta_{y_{i}}, \\ 0, \text { if } & \|\boldsymbol{z}_{i}-\boldsymbol{c}_{y_{i}}\|_{2} \leq \Delta_{y_{i}}.\end{array}\right.
	\end{equation}
	Then, we update the boundary parameter $\widehat{\Delta_{k}}$ regarding to $\mathcal{L}_{b}$ as follows:
	\begin{equation}
	\widehat{\Delta_{k}} :=\widehat{\Delta_{k}}-\eta \frac{\partial \mathcal{L}_{b}}{\partial \widehat{\Delta_{k}}},
	\end{equation} 
	where $\eta$ is the learning rate of the boundary parameters $\widehat{\Delta}$ and $\frac{\partial \mathcal{L}_{b}}{\partial \widehat{\Delta_{k}}}$ is computed by:
	\begin{equation}	
	\frac{\partial \mathcal{L}_{b}}{\partial \widehat{\Delta_{k}}}=\frac{\sum_{i=1}^{N} \delta^{'}\left(y_{i}=k\right) \cdot(-1)^{\delta_{i}}}{\sum_{i=1}^{N} \delta^{'}\left(y_{i}=k\right)}\cdot \frac{1}{1+e^{-\widehat{\Delta_{k}}}},
	\end{equation}
	where $\delta^{'}(y_{i}=k)=1$ if $y_{i}=k$ and $\delta^{'}(y_{i}=k)=0$ if not. We only update the radius $\Delta_{y_{i}}$ belonging to class $k$ in a mini-batch, which ensures the denominator is not zero. 
	
	With the boundary loss $\mathcal{L}_{b}$, the boundaries can adapt to the intent feature space and learn suitable decision boundaries. The learned decision boundaries can not only effectively surround most of the known intent samples, but also not be far away from each known class centroid, which is effective to identify the open intent samples.
	\subsection{Open Classification with Decision Boundary}
	After training, we use the centroids and the learned decision boundaries of each known class for inference. We suppose known intent samples are constrained in the closed ball area produced by their corresponding centroids and decision boundaries. On the contrary, the  open intent samples are outside any of the bounded spherical areas. Specifically, we perform open intent classification as follows:
	\begin{equation}
	\hat{y}=\left\{\begin{array}{l}
	\text {open, if } d(\boldsymbol{z}_{i},\boldsymbol{c}_{k}) > \Delta_{k}, \forall k \in \mathcal{Y}; \\
	\arg \min _{k \in \mathcal{Y}} d(\boldsymbol{z}_{i},\boldsymbol{c}_{k}), \text {otherwise},
	\end{array}\right.
	\end{equation}
	where $d(\boldsymbol{z}_{i},\boldsymbol{c}_{k})$ denotes the Euclidean distance between $\boldsymbol{z}_{i}$ and $\boldsymbol{c}_{k}$. $\mathcal{Y}=\{1,2,\cdots,K\}$ denote the known intent labels. 
	\section{Experiments}
	\begin{table*}[!t]\small
		\centering
		\begin{tabular}{@{\extracolsep{15pt}}clccccccccc}
			\toprule
			\centering
			&  & \multicolumn{2}{c}{BANKING} & \multicolumn{2}{c}{OOS} & \multicolumn{2}{c}{StackOverflow} \\
			\addlinespace[0.1cm]  \cline{3-4} \cline{5-6} \cline{7-8}  \addlinespace[0.1cm]
			& Methods &  Open & Known & Open  & Known  & Open & Known\\
			\midrule
			\multirow{6}{*}[1ex]{25\%}
			& MSP & 41.43 & 50.55 & 50.88 & 47.53   & 13.03 & 42.82 \\
			& DOC & 61.42 & 57.85  & 81.98 & 65.96  & 41.25 & 49.02 \\
			& OpenMax & 51.32 & 54.28  & 75.76 & 61.62  & 36.41 & 47.89 \\
			& DeepUnk & 70.44 & 60.88 & 87.33 & 70.73  & 49.29 & 52.60  \\									
			& ADB  & \textbf{84.56} & \textbf{70.94}  & \textbf{91.84} & \textbf{76.80}  & \textbf{90.88} & \textbf{78.82} \\
			\midrule
			\midrule
			\multirow{6}{*}[1ex]{50\%}
			& MSP & 41.19 & 71.97 & 57.62 & 70.58 & 23.99 & 66.91 \\
			& DOC & 55.14 & 73.59 & 79.00 & 78.25   & 25.44 & 66.58 \\
			& OpenMax & 54.33 & 74.76 & 81.89 & 80.54  & 45.00 & 70.49 \\
			& DeepUnk & 69.53 & 77.74  & 85.85 & 82.11  & 43.01 & 70.51  \\
			& ADB &\textbf{78.44}	&\textbf{80.96}		&\textbf{88.65}	&\textbf{85.00}		&\textbf{87.34}	&\textbf{85.68}	\\
			\midrule
			\midrule
			\multirow{6}{*}[1ex]{ 75\%}
			& MSP & 39.23 & 84.36  & 59.08 & 82.59   & 33.96 & 80.88 \\
			& DOC & 50.60 & 83.91  & 72.87 & 83.69  & 16.76 & 78.95 \\
			& OpenMax & 50.85 & 84.64 & 76.35 & 73.13   & 44.87 & 82.11 \\
			& DeepUnk & 58.54 & 84.75  & 81.15 & 86.27  & 37.59 & 81.00 \\
			& ADB &\textbf{66.47}	&\textbf{86.29}		&\textbf{83.92}	&\textbf{88.58}	 &\textbf{73.86}	&\textbf{86.80}	\\
			\bottomrule
		\end{tabular}
		\caption{ \label{results-main-2}  
			Results of open classification with different known class ratios (25\%, 50\% and 75\%) on BANKING, OOS and StackOverflow datasets. “Open” and “Known” denote the macro f1-score over open class and known classes respectively.
		}
	\end{table*}
	
	\subsection{Datasets}
	We conduct experiments on three challenging real-world datasets to evaluate our approach. The detailed statistics are shown in Table~\ref{data-stat-table}. 
	\paragraph{BANKING}
	A fine-grained dataset in the banking domain~\cite{Casanueva2020}. It contains 77 intents and 13,083 customer service queries.
	\paragraph{OOS}
	A dataset for intent classification and out-of-scope prediction~\citep{larson-etal-2019-evaluation}. It contains 150 intents, 22,500 in-domain queries and 1,200 out-of-domain queries.
	
	\paragraph{StackOverflow}
	A dataset published in Kaggle.com. It contains 3,370,528 technical question titles. We use the processed dataset~\citep{xu2015short}, which has 20 different classes and 1,000 samples for each class.
	
	\subsection{Baselines}
	We compare our method with the following state-of-the-art open classification methods: OpenMax~\citep{bendale2016towards}, MSP~\cite{hendrycks17baseline}, DOC~\cite{Shu2017DOCDO} and DeepUnk~\cite{lin-xu-2019-deep}. 
	
	As OpenMax is an open set detection method in computer vision, we adapt it for open intent classification.  We firstly use the softmax loss to train a classifier on known intents, then fit a Weibull distribution to the classifier's output logits. Finally, we recalibrate the confidence scores with the OpenMax Layer. Due to lack of open intent for tuning, we adopt default hyperparameters of OpenMax. We use the same confidence threshold (0.5) as in~\cite{lin-xu-2019-deep} for MSP. For a fairness comparison, we replace the backbone network of these methods with the same BERT model as ours.
	
	\subsection{Evaluation Metrics}
	Following previous work~\cite{Shu2017DOCDO,lin-xu-2019-deep}, we regard all the open classes as one rejected class.  To evaluate the overall performance, we use accuracy score (Accuracy) and macro F1-score (F1-score) as metrics. They are calculated over all classes (known classes and open class). We also calculate macro F1-score over known classes and open class respectively, which better evaluates the fine-grained performance.
	
	\begin{figure}
		\centering  
		\includegraphics[width=0.9\columnwidth ]{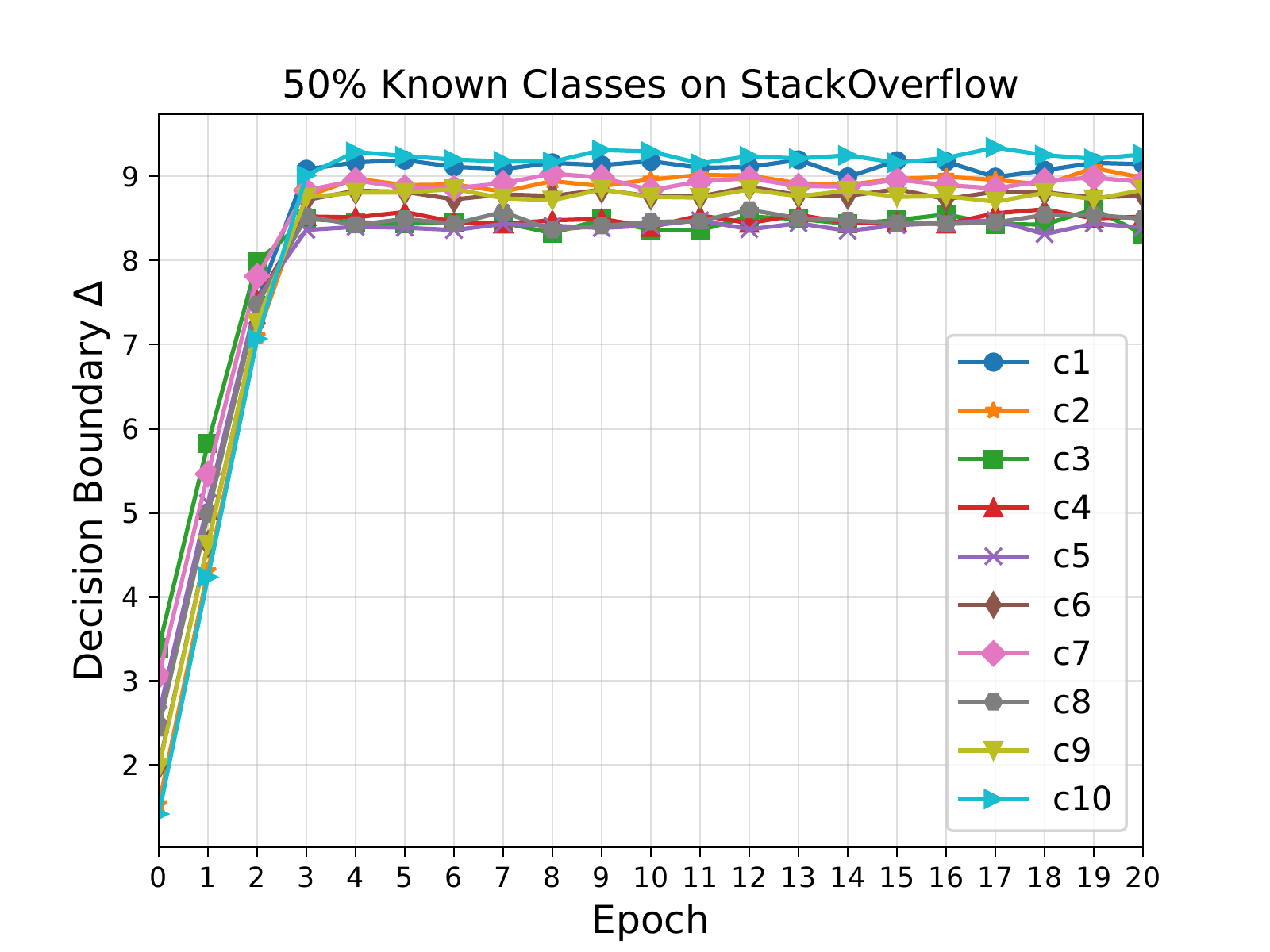}
		\caption{The boundary learning process.}
		\label{Aba_2_1}
	\end{figure}
	
	\subsection{Experimental Settings}
	Following the same settings as in ~\citep{Shu2017DOCDO,lin-xu-2019-deep}, we keep some classes as unknown (open) and integrate them back during testing. All datasets are divided into training, validation and test sets. The number of known classes are varied with the proportions of 25\%, 50\%, and 75\% in the training set. The remaining classes are regarded as one open class and removed from the training set. Both known classes and open class are used for testing. For each known class ratio, we report the average performance over ten runs of experiments.  
	
	We employ the BERT model (bert-uncased, with 12-layer transformer) implemented in PyTorch~\cite{Wolf2019HuggingFacesTS} and adopt most of its suggested hyperparameters for optimization. To speed up the training procedure and achieve better performance, we freeze all but the last transformer layer parameters of BERT. The training batch size is 128, and the learning rate is 2e-5.  For the boundary loss $\mathcal{L}_{b}$, we employ Adam~\citep{kingma2014adam} to optimize the boundary parameters at a learning rate of 0.05.
	
	\begin{figure}
		\centering  
		\includegraphics[width=0.9\columnwidth ]{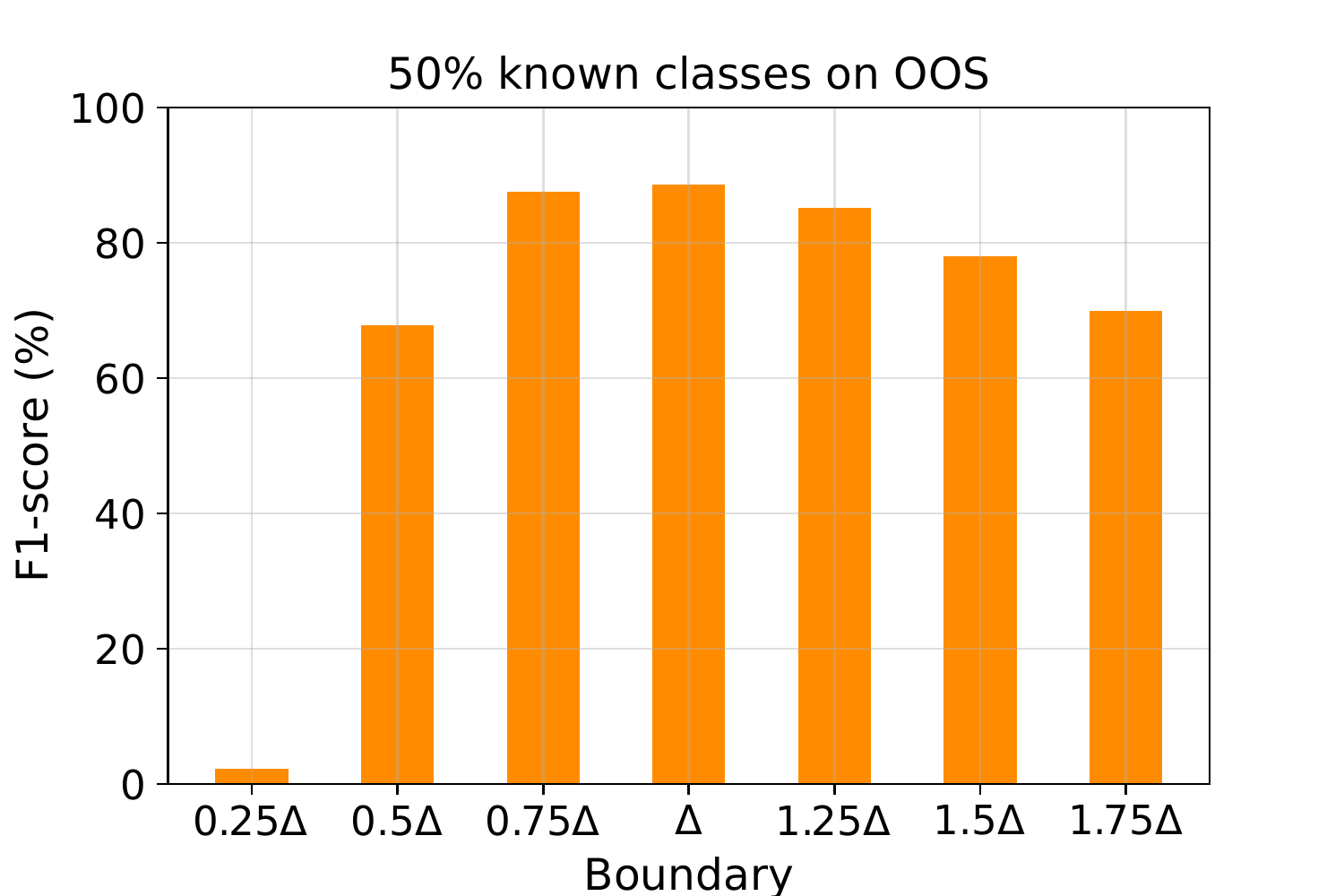}
		\caption{ Influence of the learned 	decision boundary.}
		\label{Aba_2_2}
	\end{figure}
	\begin{figure*}[!t]
		\centering  
		\includegraphics[width=2.11\columnwidth ]{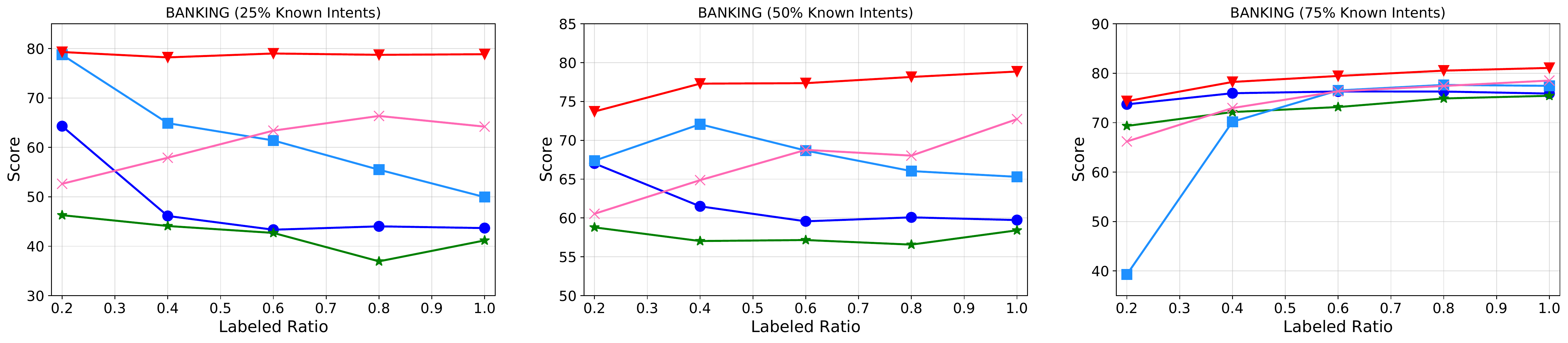}
		\includegraphics[width=2.11\columnwidth ]{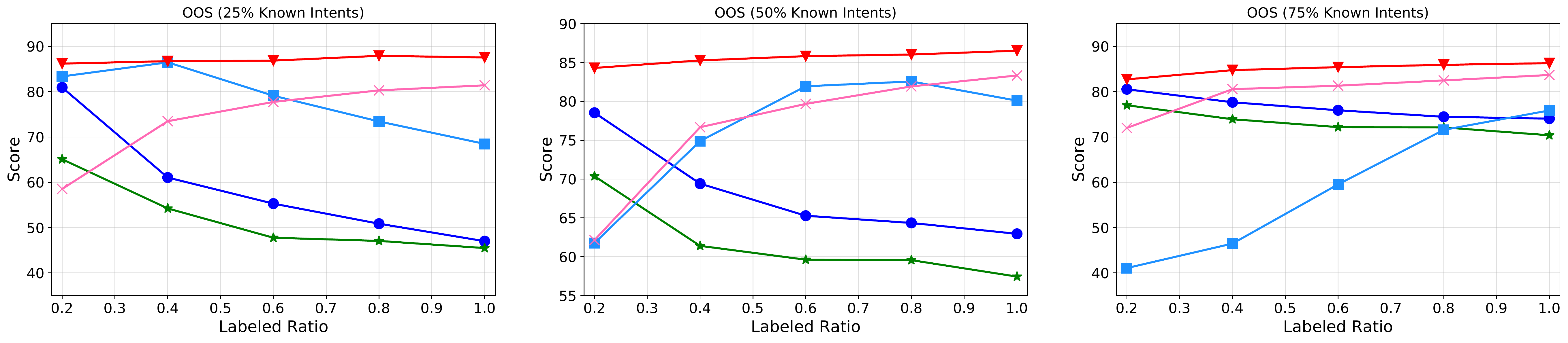}
		\includegraphics[width=2.11\columnwidth ]{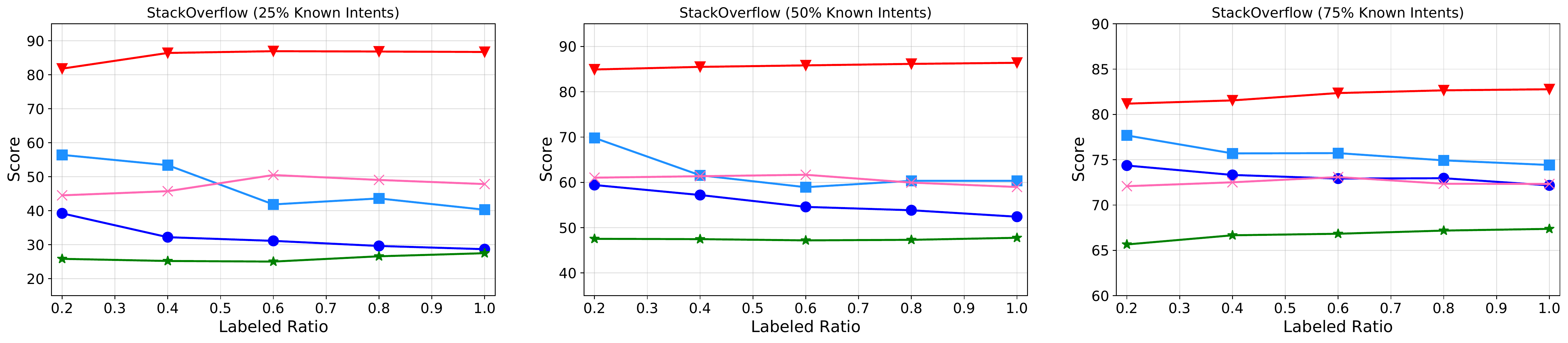}
		\includegraphics[width=2.06\columnwidth ]{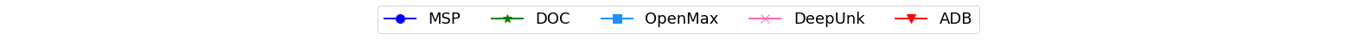}
		\caption{ Influence of the labeled ratio on three datasets with different known class proportions (25\%, 50\%, 75\%).}
		\label{Aba_3}
	\end{figure*}
	
	\subsection{Results}
	
	Table~\ref{results-main-1} and Table~\ref{results-main-2} show the performances of all compared methods, where the best results are highlighted in bold. Firstly, we observe the overall performance. Table~\ref{results-main-1} shows  accuracy score and macro F1-score over all classes. With 25\%, 50\%, and 75\% known classes, our approach consistently achieves the best results and outperforms other baselines by a significant margin. Compared with the best results of all baselines, our method improves accuracy score (Accuracy) on BANKING by 14.64\%, 6.13\%, and 2.56\%, on OOS by 6.16\%, 3.19\%, and 2.61\%, on StackOverflow by 38.88\%, 27.42\%, and 10.45\% in 25\%, 50\% and 75\% settings respectively, which demonstrates the priority of our method. 
	
	Secondly, we notice that the improvements on StackOverflow are much more drastic than the other two datasets. We suppose the improvements mainly depend on the characteristics of datasets. Most baselines lack explicit or suitable decision boundaries for identifying the open intent, so they are more sensitive to different datasets. For example, they are limited to distinguish difficult semantic intents (e.g., technical question titles in StackOverflow) without prior knowledge. By contrast, our method learns specific and tight decision boundaries for each known class, which is more effective for open intent classification. 
	
	Thirdly, we observe the fine-grained performance. Table~\ref{results-main-2} shows the macro F1-score on open intent and known intents respectively. We notice that our method not only achieves substantial improvements on open class, but also largely enhances the performances on known classes compared with baselines. That is because our method can learn specific and tight decision boundaries for detecting open class while ensuring the quality of known intent classification. 
	
	\section{Discussion}
	
	\subsection{Boundary Learning Process}
	Figure~\ref{Aba_2_1} shows the decision boundary learning process. At first, most parameters are assigned small values near zero after initialization, which leads to small radius with the $\operatorname{Softplus} $ activation function. As the initial radius is too small, the empirical risk plays a dominant role. Therefore, the radius of each decision boundary expands to contain more known intent samples belonging to its class. As the training process goes on, the radius of the decision boundary learns to be large enough to contain most of the known intents. However, the large radius will also introduce redundant open intent samples. In this case, the open space risk plays a dominant role, which prevents the radius from enlarging. Finally, the decision boundaries converge with a balance between empirical risk and open space risk.
	
	\subsection{Effect of Decision Boundary}
	To verify the effectiveness of the learned decision boundary, we use different ratios of $\Delta$ as boundaries during testing. As shown in Figure~\ref{Aba_2_2}, ADB achieves the best performance with $\Delta$ among all assigned decision boundaries, which verifies the tightness of the learned decision boundary. Moreover, we notice that the performance of open classification is sensitive to the size of the decision boundaries. Overcompact decision boundaries will increase the open space risk by misclassifying more known intent samples to the open intent. Correspondingly, overrelaxed decision boundaries will increase the empirical risk by misclassifying more open intent samples as known intents. As shown in Figure~\ref{Aba_2_2}, both of these two cases perform worse compared with $\Delta$. 
	
	\subsection{Effect of Labeled Data}
	To investigate the influence of labeled data, we vary the labeled ratio in the training set in the range of 0.2, 0.4, 0.6, 0.8 and 1.0. We use Accuracy as the score to evaluate the performance. As shown in Figure~\ref{Aba_3}, ADB outperforms all the other baselines on three datasets on almost all settings. Besides, it keeps a more robust performance under different labeled ratios compared with other methods. 
	
	Notably, the statistic-based methods (e.g., MSP and DOC) show better performances with less labeled data. We suppose the reason is that the predicted scores are in low-confidence with less prior knowledge for training, which is helpful to reject the open intent with the threshold. However, as the number of labeled data increases, these methods tend to be biased towards the known intents, with the aid of strong feature extraction capability of DNNs~\cite{7298640}. Therefore, the performances drop dramatically.

	In addition, we notice that OpenMax and DeepUnk are two competitive baselines. We suppose the reason is that they both leverage the characteristics of intent feature distribution to detect the open class. However, OpenMax computes centroids of each known class with only corrective positive training samples. The qualities of centroids are easily influenced by the number of training samples. DeepUnk adopts a density-based novelty detection algorithm to perform open classification, which is also limited to the prior knowledge of labeled data. Thus, their performances all drop dramatically with less labeled data, as shown in Figure~\ref{Aba_3}.

	\subsection{Effect of Known Classes}
	We vary the known class ratio between 25\%, 50\% and 75\%, and show the results in Table~\ref{results-main-1} and Table~\ref{results-main-2}. Firstly, we observe the overall performance in Table~\ref{results-main-1}. Compared with other methods, our method achieves huge improvements on all settings of three datasets. All baselines drop dramatically as the known class ratio decreases. By contrast, our method still achieves robust results on accuracy score with fewer training samples. 
	
	Then, we observe the fine-grained performance in Table~\ref{results-main-2}. We notice that all baselines achieve high scores on known classes, but they are limited to identify the open intent and suffer poor performance. However, our method still yields the best results on both known classes and the open class. It further demonstrates that the suitable learned decision boundaries are helpful to both balance the empirical risk and the open space risk.
	
	\section{Related Work}
	\subsection{Intent Detection} 
	There are many works for intent detection in dialogue systems in recent years~\cite{ijcai2020-532,Qin_Che_Li_Ni_Liu_2020,zhang-etal-2019-joint,e-etal-2019-novel,qin-etal-2019-stack}. Nevertheless, they all make assumptions of closed world classification without the open intent.~\citet{srivastava-etal-2018-zero} perform intent detection with the zero-shot learning (ZSL) method. However, ZSL is different from our task because it only contains novel classes during testing. 
	
	Unknown intent detection is a specific task for detecting the unknown intent.~\citet{Brychcin2017UnsupervisedDA} propose an unsupervised approach to modeling intents,  but fail to utilize the prior knowledge of known intents.~\citet{Kim2018JointLO} jointly train the in-domain (ID) classifier and out-of-domain (OOD) detector but need to sample OOD utterances.~\citet{Yu2017OpenCategoryCB} adopt adversarial learning to generate positive and negative samples for training the classifier.~\citet{ryu-etal-2018-domain} use a generative adversarial network (GAN) to train on the ID samples and detect the OOD samples with the discriminator. However, it has been shown that deep generative models fail to capture high-level semantics on real-world data~\cite{2018arXiv181009136N,Mundt_2019_ICCV}. Recent methods try to learn friendly features for detecting the unknown intent ~\cite{lin-xu-2019-deep,gangal2020likelihood,yan-etal-2020-unknown}, but they need to modify the  model architecture, and fail to construct specific decision boundaries.
	
	\subsection{Open World Classification}
	At first, researchers use SVM to solve open set problems. One-class classifiers~\cite{scholkopf2001estimating,SVDD} find the decision boundary based on the positive training data. For multi-class open classification, One-vs-all SVM~\cite{Rifkin2004In} trains the binary classifier for each class and treats the negative classified samples as the open class.~\citet{scheirer2013toward} extend the method to computer vision and introduce the concept of open space risk.~\citet{Jain_2014_ECCV} estimate the unnormalized posterior probability of inclusion for open set problems. They fit the probability distributions to statistical Extreme Value Theory (EVT) by using a Weibull-calibrated multi-class SVM.~\citet{Scheirer_2014_TPAMIb} propose a Compact Abating Probability (CAP) model, which further improves the performance of Weibull-calibrated SVM by truncating the abating probability. However, all these methods need negative samples for selecting the decision boundary or probability threshold. Moreover, SVM cannot capture  advanced semantic features of intents~\cite{lin2019post}.
	
	Recently, researchers use deep neural networks for open classification. OpenMax~\cite{bendale2016towards} fits Weibull distribution to the outputs of the penultimate layer, but still needs negative samples for selecting the best hyperparameters. MSP~\cite{hendrycks17baseline} calculates the softmax probability of known samples and rejects the low confidence unknown samples with the threshold. ODIN~\cite{liang2018enhancing} uses temperature scaling and input preprocessing to enlarge the differences between known and unknown samples. However, both of them~\cite{hendrycks17baseline,liang2018enhancing} need unknown samples to select the confidence threshold  artificially. DOC~\cite{Shu2017DOCDO} uses the sigmoid function and calculates the confidence threshold based on Gaussian statistics, but it performs worse when the output probabilities are not discriminative. 
	
	\section{Conclusion}
	In this paper, we propose a novel post-processing method for open intent classification. After pre-training the model with labeled samples, our model can automatically learn specific and tight decision boundaries adaptive to the known intent feature space. Our method has no require for open intent or model architecture modification. Extensive experiments on three benchmark datasets show that our method yields significant improvements over the compared baselines, and is more robust with less labeled data and fewer known intents.
	
	\section{Acknowledgments}
	This work is supported by seed fund of Tsinghua University (Department of Computer Science and Technology)-Siemens Ltd., China Joint Research Center for Industrial Intelligence and Internet of Things. 
	
	\bibliography{aaai21.bib}
\end{document}